# Efficient Method for Categorize Animals in the Wild

Abulikemu Abuduweili
School of EECS,
Peking University
abuduwali@pku.edu.cn

Xin Wu
School of EECS,
Peking University
blesswoo@pku.edu.cn

Xingchen Tao
School of EECS,
Peking University
taoxingchen@pku.edu.cn

## Abstract

*Automatic species classification in camera traps would greatly help the biodiversity monitoring and species analysis in the earth. In order to accelerate the development of automatic species classification task, "Microsoft AI for Earth" have prepared a challenge in FGVC6 workshop at CVPR 2019, which called "iWildCam 2019 competition"[1]. In this work, we propose the efficient method for categorize animals in the wild. We transfer the state-of-the-art ImagaNet pretrained models to the problem. To improve the generalization and robustness of the model, we utilize efficient image augmentation and regularization strategies, like cutout, mixup and label-smoothing. Finally, we use ensemble learning to increase the performance of the model. Thanks to advanced regularization strategies and ensemble learning, we got top 7/336 places in final leaderboard. Source code of this work is available at https://github.com/Walleclipse/iWildCam_2019_FGVC6*

## 1. Introduction

The iWildCam Challenge 2019 is a fine-grained image classification competition, which is a part of FGVC6 workshop at CVPR 2019. What makes it different from all the other image classification tasks is that all of the images are collected by camera traps, which are wildly used by biologists all over the world to monitor biodiversity and population density of animal species. The problem of identification and classification of animals in wildlife footage manually is a tedious and time consuming task. However, designing an automatic system for classification of animals is a very effortful job because the images captured in real-time involve animals with complex backgrounds, different postures and different illuminations.

### 1.1. Challenges

In the camera traps animal classification task, the animals in the images can be challenging to detect, even for humans. We summarize two main challenges as follows.

**Challenges on categories:** 1) The training data is from The American Southwest, while the test data are from the American Northwest. There are data drift or concept drift between training set and testing set, we need to recognizing animals in new regions. 2) The species seen in each region overlap but are not identical. The total number of categories is 23，but training data only contain 14 categories. There is open-set problem of recognizing species of animals that have never before been seen. In addition, the training data is very unbalance to each categories. There are more than 60% images are non-animal variability as shown in Fig. 1.

**Challenges on images:** Camera trap data provides several challenges that can make it difficult to achieve accurate results. We show some challenges on images as Fig.2. 1) Illumination. Images can be poorly illuminated, especially as night. Thus make it difficult even for human to categorize the animal. For example, there is a skunk to the center left of the frame. 2) Motion blur. Some animals are too fast to be captured by the shutter of the camera. The following example contains a blurred coyote. 3) Small ROI. Some animals are small or far from the camera. The example image has a mouse on a branch to the center right of the frame. Honestly, I can't distinguish anything from a bunch of interlacing branches. 4) Occlusion. Animals can be occluded by vegetation or the edge of the frame. This example shows a location where weeds grew in front of the camera. 5) Perspective. Some animals can be very close to the camera. 6) Weather Conditions. Poor weather, including rain, snow, or dust, can obstruct the lens and cause false triggers. 7) Camera Malfunctions. Sometimes the camera malfunctions, causing strange discolorations. 8) Temporal Changes. At any given location, the background changes over time as the seasons change. Below, you can see a single location at three different points in time. 9) Non-Animal Variability. What causes the non-animal images to trigger varies based on location. Some locations contain lots of vegetation, which can cause false triggers as it moves in the wind. Others are near roadways, so can be triggered by cars or bikers.

[1] https://www.kaggle.com/c/iwildcam-2019-fgvc6

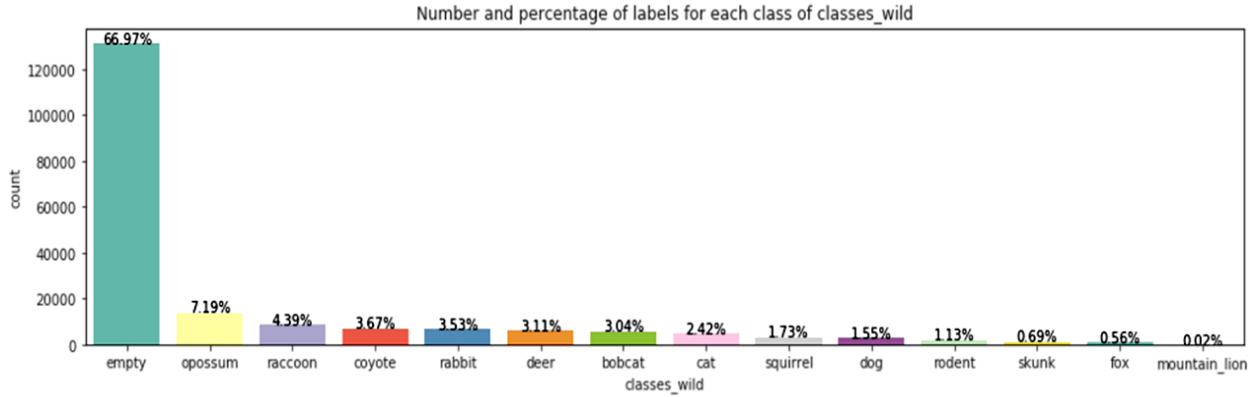

Fig. 1. Number and percentage of labels for each class

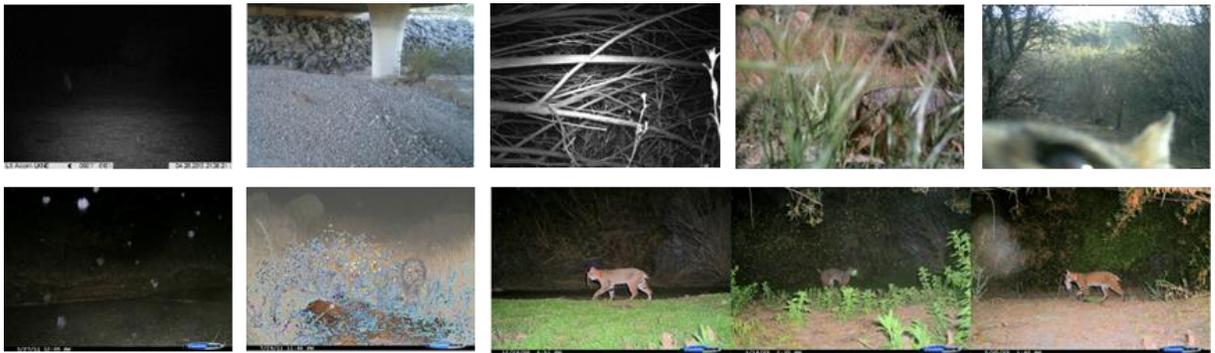

Fig. 2. challenges on image

## 1.2. Related Work

Camera traps are a valuable tool for studying biodiversity, but research using this data is limited by the speed of human annotation. With the vast amounts of data now available it is imperative that we develop automatic solutions for annotating camera trap data in order to allow this research to scale. A promising approach for camera trap animal classification is based on deep networks trained on human-annotated images [2]. Beery S et al. provide a challenge dataset to explore whether such solutions generalize to novel locations, since systems that are trained once and may be deployed to operate automatically in new locations would be most useful [1].

In recent years, a handful of ecologists has begun utilizing deep learning systems for species and animal individual identification and classification with great success. In 2014, Carter et al. published one of the first works using neural networks for animal monitoring in the wild [2]. Carter et al. 's work has been considered a large success and is currently used to monitor the southern Great Barrier Reef green turtle population. In 2017, Brust et al. trained the object detection method YOLO to extract cropped images of Gorilla faces from 2,500 annotated camera trap images of 482 individuals taken in the Western Lowlands of the Nouabal′e -Nodki National Park in the Republic of Congo [4]. Once the faces are extracted, Brust et al. train the CNN AlexNet achieving a 90.8% accuracy on a test size of 500 images. The authors close discussing how deep learning for ecological studies show promises for a whole realm of new applications if the fields of basic identify, spatio-temporal coverage and socioecological insights. In 2018, Koerschens considered a completed pipeline of object detector and classifier to re-identify elephant individuals considering a dataset from The Elephant Listening Project from the DzangaSangha reserve of which 2,078 images of 276 different individuals are present [5].

## 1.3. Contribution

In this paper, we propose the efficient method for classifying animals in the wild. We finetune the state-of-the-art ImagaNet pretrained models to suitable for this task. To improve the generalization and robustness of the model, we utilize efficient image augmentation and regularization strategies, including cutout, mixup and label-smoothing. Finally, we use ensemble learning to increase the performance of the model. Thanks to advanced regularization strategies and ensemble learning, we got top 3% (7/336) places in final leaderboard.

The rest of the paper is organized as follows. Section 2 introduces our method with relevant background knowledge. Section 3 describes the experiment and results of the competition. Section 4 demonstrate our other attempts for this competition. The last section is the summary and prospect of the work.

## 2. Method

We propose detection and classification two step strategy for this task. The key parts of our method include animal detection, image augmentation, fine-tuning of pretrained model, advanced regularization strategy and ensemble learning. Which is elaborated on in detail as follows.

### 2.1. Animal Detection with Faster-RCNN

Detecting moving objects from the background is an important and enabling step in intelligent video analysis. For camera-traps image sequences, handle highly dynamic background scenes is one of the challenges. As described in introduction, some animals are small or far from the camera, that makes very small ROI for animal area. Effective animal detection is crucial for further classification task. In our method, we implement animal detection using Faster-RCNN [6], then crop the detected animal area of the image for further classification task, as Fig.3

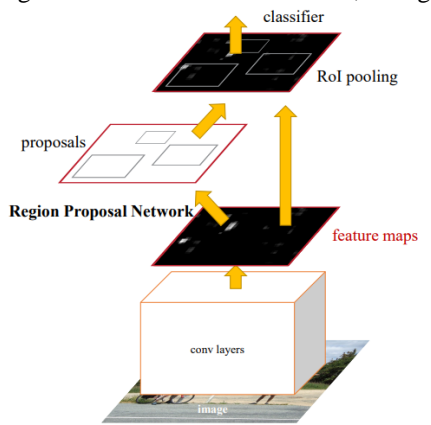

Fig.3 Faster-RCNN framework

### 2.2. Data Augmentation

Camera trap data classification task difficult to achieve accurate results, because of the quality of images. In order to make the model learn the "real animals" rather than its background, we use a lot of image augmentation strategies. Which can force the model learn the difficult cases in training step and improve the classifier results in testing step. We use following three type of image augmentation strategies.

**Traditional image transform:** For increase the rotation, scaling and anti-noise robustness for the model, we use some image transformation in training procedure, including random cropping, random rotation, horizontal flip, random brightness, random blur, adding gaussian noise, As Fig. 4

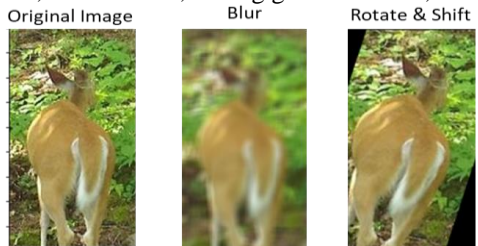

Fig. 4. image-augmentation

**CLAHE:** Contrast Limited Adaptive Histogram Equalization (CLAHE) [7] is a computer image processing technique used to improve contrast in images. It is suitable for improving the local contrast and enhancing the definitions of edges in each region of an image, As Fig. 5. In CLAHE, the contrast amplification in the vicinity of a given pixel value is given by the slope of the transformation function. This is proportional to the slope of the neighbourhood cumulative distribution function (CDF) and therefore to the value of the histogram at that pixel value. CLAHE limits the amplification by clipping the histogram at a predefined value before computing the CDF.

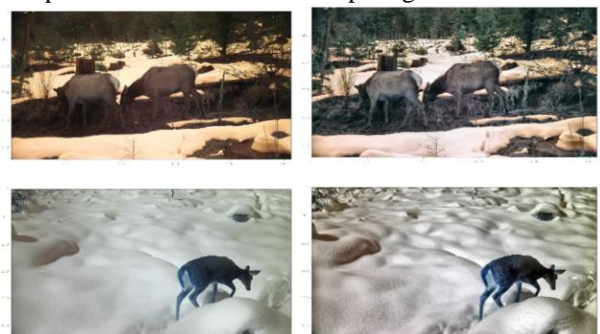

Fig. 5.Original Image(left), image with CLAHE( right)

**Grayscale:** Some of camera trap wild animals captured at dark night. This makes animal pictures lose RGB color information. In order to increase the model robustness of day and night animal pictures, we randomly grayscale the images for training, which force the model pay more attention in shape of animals rather than colors, as Fig. 6.

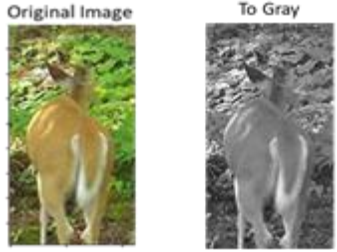

Fig. 6. image-to-gray

## 2.3. Classifier Modeling

In recent years, deep learning has exploded over the past few years thanks to big advances in machine processing power (gpus), massive amounts of data (Imagenet) and advanced algorithms. Modern deep neural networks have shown great success learning the necessary features for image classification from data and remove the need for feature engineering. We introduce some success networks for image classification as follows.

**VGGNet:** It created by Simonyan et al. from Oxford University in 2014 [8], came second in the ISLVRC 2014. VGGNet is winning solution for ImageNet Challenge 2014 localization track and it is very powerful for extract features from images. VGGNet contains 16-19 layers of weighted networks, which are deeper and more numerous than previous network architectures, As Fig. 7. The success of VGGNet shows how you can improve network performance by adding layers and depth to previous network architectures.

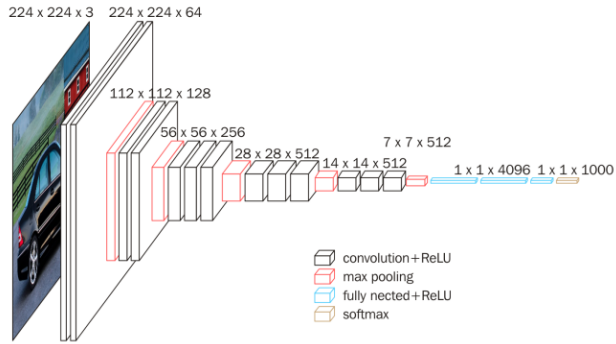

Fig. 7. VGGNet structure

**ResNet:** In 2015, Kaiming He et al. proposed ResNet [9] and won the 1st places on the tasks of ImageNet detection, ImageNet localization, COCO detection, and COCO segmentation. ResNet address the degradation problem by utilizing a deep residual learning framework. Which is a network learning framework of shortcut connection that is deeper than the previous network, as Fig.8. The advantage of this network is that it is easier to optimize and can bring significant accuracy improvement from increasing the number of network layers.

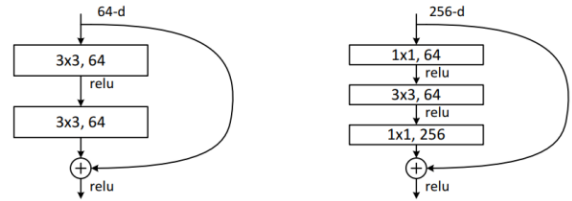

Fig. 8. Resnet shortcut connection

**DenseNet**: Gao Huang et al. proposed the concept of DenseCNN in 2016 [10], which links each layer with other layers in the feedforward process. For each layer of network, the feature maps of all the previous networks are used as input, and their feature maps are also used by other network layers as input. Each layer has direct access to the gradients from the loss function and the original input signal, leading to an implicit deep supervision, as Fig.9. This helps training of deeper network architectures.

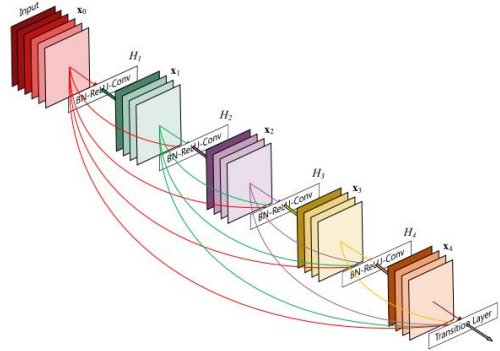

Fig. 9. Densnet structure

**EfficientNet:** In 2019, Mingxing Tan et al. investigate the scaling up ConvNets and propose the state-of-the-art architecture EfficientNet [11]. Mingxing Tan et al. used neural architecture technique to obtain efficient baseline network, which called EfficientNet-B0. Then apply depth, width and resolution compound scaling to obtain more effective models EfficienNet-B1 ~ EfficientNet-B7, as Fig.10. Which got the state-of-the art performance in ImageNet dataset until the June 2019.

## 2.4. Advanced Regularization Strategy

In iWildcam 2019, the training data and the testing data is from different regions. Deep CNN networks is very easy to overfit the training set and related regions. To handle the overfitting problem in image classification task, a lot of advanced regulation strategies emerges. We introduce the effective regularization strategies as follows.

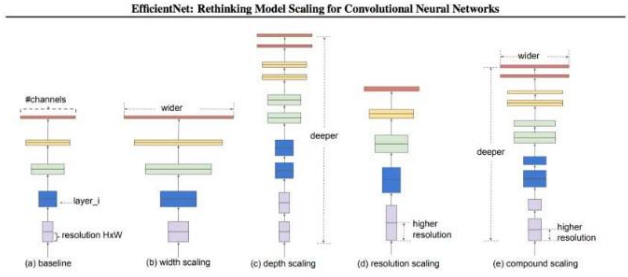

Fig.10 Model scaling for EfficientNet

**Label smoothing:** Label smoothing is a mechanism of regularize the classifier layer by estimating the marginalized effect of label-dropout during training [12]. It can effectively suppress the over-fitting phenomenon when calculating the loss value by "softening" the traditional one-hot type label as following function:

$$q'(k) = (1-\epsilon)\delta_{k,y} + \frac{\epsilon}{K}$$

$K$ is the number of classes, $\varepsilon$ is a hyperparameter, usually set as 0.1. Thus, label smoothing is equivalent to replacing a single cross-entropy loss $H(q, p)$ with a pair of such losses $H(q, p)$ and $H(u, p)$.

$$H(q',p) = -\sum_{k=1}^{K}\log p(k)q'(k) = (1-\epsilon)H(q,p)+\epsilon H(u,p)$$

**Cutout:** Cutout is a simple regularization technique of randomly masking out square regions of input during training, can be used to improve the robustness and to avoid overfitting [13], as Fig.11 . Which involves removing contiguous sections of input images, effectively augmenting the dataset with partially occluded versions of existing samples. This technique can be interpreted as an extension of dropout in input space, but with a spatial prior applied, much in the same way that CNNs apply a spatial prior to achieve improved performance over feed-forward networks on image data.

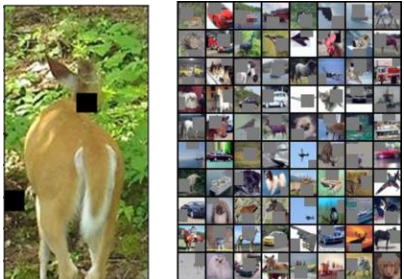

Fig. 11. Cutout

**Mixup:** Mixup is a data-agnostic and straightforward data augmentation principle [14]. Mixup is a form of vicinal risk minimization, which trains on virtual examples constructed as the linear interpolation of two random examples from the training set and their labels. In a nutshell, sampling from the mixup vicinal distribution produces virtual feature-target vectors

$$\tilde{x} = \lambda x_i + (1-\lambda)x_j,$$
$$\tilde{y} = \lambda y_i + (1-\lambda)y_j,$$

where $(x_i , y_i)$ and $(x_j , y_j )$ are two feature-target vectors drawn at random from the training data, and $\lambda \in [0, 1]$. Mixup improves the generalization error of state-of-the-art models on ImageNet, CIFAR, speech, and tabular datasets.

### 2.5. Ensemble Learning

Experimental evidence has shown that ensemble methods are often much more accurate than single hypothesis [15]. Learning algorithm that output only a single hypothesis suffer problems that can be partly overcome by ensemble methods: the statistical problem, the computational problem and the representation problem.

We use weighted average strategy in this task. We have 3 models Renset 101, EfficientNet-B0 and EfficientNet-B3. For each models we predict the test images with 3 different preprocessing methods, that are original image, CLAHE and grayscale. For sum up, we have 9 prediction results, which is 3 models prediction × 3 test data preprocessing. We averaging the output probability of 9 prediction results and got better performance than any single models.

### 3. Experiment and Results

**Implementation:** For the final submission of the iWildCam 2019 competition, we use Resnet-101, EfficientNet-B0 and EfficientNet-B3 as our base model, and submit the ensemble results for above three models. We used Pytorch framework for implementation. For the training, we used Adam optimizer with batch size set to 32 and training 20 epochs. We used multistep learning rate decay strategy for effective training. The initial value of learning rate is set to 0.005, and learning rate decays with the factor of 0.1 at $2^{nd}$ and $5^{th}$ epoch.

**Metrics:** The major evaluation metrics is macro F1 score. F1 will be calculated for each class of animal (including "empty" if no animal is present), and the final score will be the unweighted mean of all class F1 scores.

**Results:** The results are shown in Table 1. We used EfficientNet-B0, EfficientNet-B3 and ResNet-101. For each model, we used 3 different test image preprocessing methods,. As can be seen, ensemble result obtain the best performance on Macro-F1 score. For the single model, EfficienNet-B0 with test image CLAHE achieved the best single model results.

Table 1. Final Leaderboard score for iWildCam 2019

| models | Preprocessing | Macro-F1 |
|---|---|---|
| EfficientNet-B0 | Original | 0.224 |
|  | CLAHE | 0.225 |
|  | Grayscale | 0.210 |
| EfficientNet-B3 | Original | 0.201 |
|  | CLAHE | 0.200 |
|  | Grayscale | 0.159 |
| ResNet-101 | Original | 0.169 |
|  | CLAHE | 0.179 |
|  | Grayscale | 0.173 |
| **Ensemble** | **-** | **0.228** |

## 4. Some Attempts in Competitions

Besides the methods we mentioned above, we also investigate the other solutions for the competition. Although these attempts perform not very good, we think these attempts are worth discussing

### 4.1. Classification Rare Species Individually

Instead of augmenting the training data in order to make our model as robust as possible to the test data that has an unknown distribution and assembling different models, compensating for each other's shortcomings, we quit data augmentation and tried to organize different models vertically as Fig. 12.

Considering the high imbalance within the distribution of the training data, firstly we randomly picked 3000 to 4000 (the exact number can be a hyper-parameter to be tuned) samples from those classes having samples more than the number, and then we chose all samples of the classes with between 2000 to 3000 ones. The tricky part came to the rest of the categories, some of which can have fewer than 10 samples. We grouped those categories into new ones according to the resemblance in shape and the maintenance of the balance of the data distribution. There were 2 new categories as a result, consisting of 3 and 5 original ones respectively. These new categories will be called Rare Species 1 and Rare Species 2 in the following. (Rare Species 1 is comprised of deer, moose and elk. Moose and elk are basically the same species, except for the fact that one inhabits North America and the other Europe. They can be regarded as deer with a larger size. Rare Species 2 consists of pronghorn antelope, bighorn sheep, bison, mountain sheep and mountain lion. These 5 categories may not share as much resemblance in shape as those do in Rare Species 1. The reason why they were grouped together is that they all are classes with around 300 samples.)

Next, we trained a VGG16 on these 17 (23-8+2=17) classes. A Relation Network was trained on the original 23 classes in the meantime. The relationship between the two models is depicted as follows:

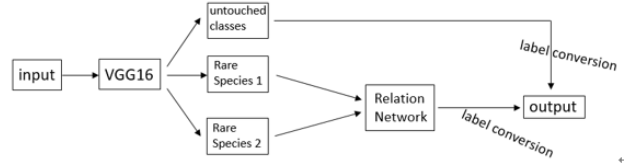

Fig. 12. Pipelined on rare species

If an input image is categorized as one of the 'rare species', that image is further processed by the Relation Network. Number of the support classes are 3 and 5 respectively, depending on which category the input image is categorized as by the VGG16. In practice, we set the number of images per class in the support set to be 15. The label has be converted back to the one that corresponds to the original class set before it is output.

To our surprise, no prediction was 'empty' during inference. We conjecture it's because there are too many kinds of 'things' that an empty image could be. So we modified the above architecture a little bit as Fig.13.

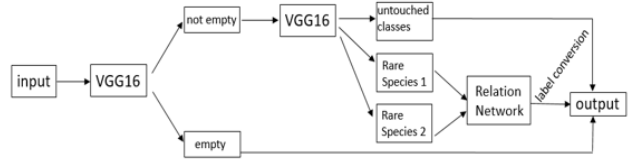

Fig. 13. Pipeliend on not empty species

We trained a second VGG16, responsible for judging if an image is empty. Now because there are only 2 classes, the number of samples per class can rapidly increase from 3000-4000 to 50000-60000. This architecture resulted in a marco-F1 of 0.154. Worse than our best result. Due to the time limit, we didn't do any ablation test. Because of the wild range of diversity in the background (different animals can be captured in the same background, and the same kind of animal can be captured by different cameras deployed in different locations.), data augmentation is very likely to play an important role in the final result, we guess.

### 4.2. Classification Original Image with Data Balancing

**Data balancing:** As mentioned before, the training data is very unbalance to each categories. There are more than 60% images are non-animal variability. Data distribution may influence model effect, data balancing is usually used to avoid overfitting on dataset.

**Classify the original image without detection:** Animal detection may misses the some hard detecting animals with low image quality. If we classify the original image without detection, we can skip the errors caused by detection procedure. We investigate the original image classification.

Due to the color change or night shooting occurs in large number of images, we improve brightness and contrast by CLAHE, and correct the color by white balance. We Finetune the ImageNet pretrained models of DenseNet121, VGG16, VGG19, Xception.

Ensemble results of original image classification (without detection) are shown in Table 2. As can be seen, data balancing reduce the performance of the model, that is because balancing training data may increase the difference of distributions between training and testing set. By the comparison of Table 1, classification in original image have a lower performance rather than detection and classification 2 step strategy. So for the final submission, we used detection and classification two - step strategy without data balancing.

Table 2. Classifying original image (without detection)

| models | Preprocessing | Macro-F1 |
|---|---|---|
| Ensemble | Data balancing | 0.121 |
| | Without balancing | 0.132 |

## 5. Conclusion

The iWildCam Challenge 2019 is a camera traps wild animal classification tasks. The task help the biologists all over the world use camera traps to monitor biodiversity and population density of animal species. In this paper, we propose the efficient method for categorize animals in the wild. We transfer the ImageNet pretrained models to iWildCam challenges. We implement image augmentation (image transformation, CLAHE, grayscale) and regularization strategies (cutout, mixup and label-smoothing) to improve the generalization and robustness of the model. We got top 3% (7/336) places in final leaderboard.

There is still improvement potential for our approach. As we mentioned before, the training data and test data are from different regions, which cause domain shift between two dataset. In the future, we will investigate the domain shift problem between training set and testing set. Generative Adversarial Nets (GAN) may be a potential solution.